# Autonomous Resource Management in Construction Companies Using Deep Reinforcement Learning Based on IoT⋆

Maryam Soleymani[a,∗], Mahdi Bonyani[b] and Meghdad Attarzadeh[c,d]

[a]Department of Construction and Project Management – Art University of Tehran, Tehran, Iran
[b]Department of Computer Engineering, University of Tabriz, Iran
[c]Department of Construction Management, School of Architecture & Planning, Morgan State University, USA
[d]Department of Engineering Technology, Old Dominion University, USA

ABSTRACT

Resource allocation is one of the most critical issues in planning construction projects, due to its direct impact on cost, time, and quality. There are usually specific allocation methods for autonomous resource management according to the project's objectives. However, integrated planning and optimization of utilizing resources in an entire construction organization are scarce. The purpose of this study is to present an automatic resource allocation structure for construction companies based on Deep Reinforcement Learning (DRL), which can be used in various situations. In this structure, Data Harvesting (DH) gathers resource information from the distributed Internet of Things (IoT) sensor devices all over the company's projects to be employed in the autonomous resource management approach. Then, Coverage Resources Allocation (CRA) is compared to the information obtained from DH in which the Autonomous Resource Management (ARM) determines the project of interest. Likewise, Double Deep Q-Networks (DDQNs) with similar models are trained on two distinct assignment situations based on structured resource information of the company to balance objectives with resource constraints. The suggested technique in this paper can efficiently adjust to large resource management systems by combining portfolio information with adopted individual project information. Also, the effects of important information processing parameters on resource allocation performance are analyzed in detail. Moreover, the results of the generalizability of management approaches are presented, indicating no need for additional training when the variables of situations change.

## 1. Introduction

In today's highly competitive construction industry, resource management is an important management tool that contributes to improving the performance of construction companies with several projects [1, 2]. Indeed, it is possible to maximize the company's profit by allocating limited resources rationally among competitive objectives, based on the rule of effectiveness [3]. Additionally, appropriate allocation of resources in construction projects plays a critical role in the planning process as it directly impacts cost, duration, and quality [4]. The first challenge is to allocate resources to different construction processes according to what is needed, when and where it is needed, and what is available [5]. In most classical planning methods, the project plan is developed at the planning phase and it is expected to be executed according to that plan, regardless of any possible changes during the project implementation [4]. However, there are numerous methods used for the optimization of resource allocation and leveling, giving better results compared to traditional methods [2].In fact, resources must be leveled in the project to avoid difficulties in the construction works due to variations in resource use [6].

Research on optimization of the use of resources by construction companies primarily involves methods of scheduling individual construction projects, rather than a portfolio, taking into account limitations in resource availability. In the meantime, designing an optimum schedule including resource planning for a single building structure (e.g., with a minimum construction cycle) and its implementation according to the schedule do not guarantee a construction company's efficiency. On the other hand, the role of the management staff of a construction company is to maintain balance between the production capacity of a company and the portfolio of orders [7, 8]. In this way, computerized decision-support systems and optimizing methods can enhance the quality of management decisions [9, 10]. Using a computer-based resource management system, construction companies can keep track of daily updates on their resources, make corrections based on the previous progress of their work on all sites, and update their goals weekly. The purpose is to implement tasks that will help to stick to deadlines for completing construction stages [3, 10].

Autonomous resource management (ARM) applies efficient and safe resource allocation methods regarding resource constraints and objectives such as time [11, 12]. These methods include coverage resource allocation (CRA) [13, 14] for the projects and data harvesting (DH) from Internet of Things (IoT) nodes [15, 16]. Typically, CRA is used to manage resources between the start and end points of the projects, with the goal of covering all points' resource allocation in the project of interest. As far as possible, CRA covers the target project considering obstructions such as constraints in resources [17].

In the construction company, DH scenario involves ARM

∗Corresponding author
✉ m.soleymani.pm@gmail.com (M. Soleymani)
ORCID(s): 0000-0003-3796-3137 (M. Soleymani); 0000-0003-0922-9656 (M. Bonyani)

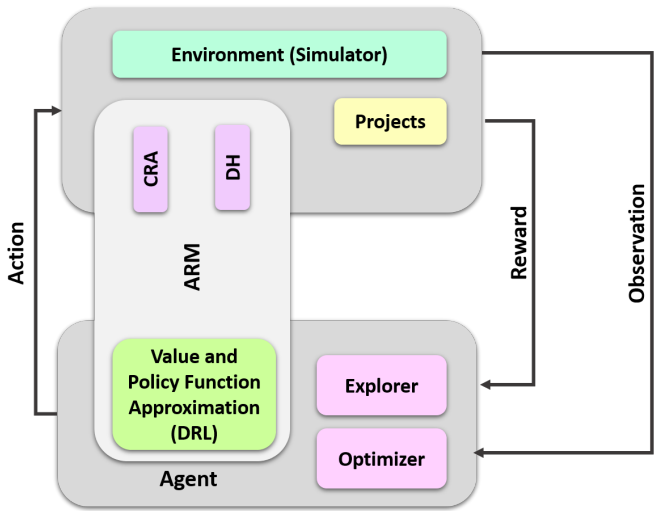

**Figure 1:** Agent construction company interactions for ARM problem.

collecting data from IoT devices distributed throughout the projects' sites, as is shown in Fig. 1, through an alternating line-of-sight (LoS) and non-line-of-sight (NLoS) link [18]. There are many similarities between a DH problem and a CRA problem when presented as an RL problem, since both have very similar constraint sets. The only major difference is the goal function. In previous researches, CRA and DH have been studied separately. They show that both problems are solvable with deep reinforcement learning (DRL) by putting spatial data of the construction project's company simulation into the DRL agent via layers of convolutional networks [16, 19, 20, 21, 22, 23, 24, 25]. In this work, we propose improvements to existing DRL approaches through generalizing ARM using the methods of CRA and DH in construction companies.

Large amount of portfolio information makes it problematic to use simulation maps as a direct input because the network's size, training parameters, and training time increase [26]. In fact, scalability issues presented by the standard simulation map input are addressed by portfolio and individual project information. A shortage in the resource feature inhibits taking the required actions promptly whereas focus on start time feature tends to the lack of sufficient resources in making decision. As a result, start time can provide less detailed information to the agent than resources shortages. An agent is provided with general information about all resources on the simulation map of the company's portfolio and individual project. An individual project simulation provides detailed local information that is uncompressed leading to immediate but unreliable action at surroundings of the ARM agent. The DRL provides the possibility of solving both distinctly different problems, namely DH and CRA problems, with the same approach, despite the fact that numerous resource allocation algorithms exist for each problem [27, 28, 29, 30]. As a result, DRL agent learning the management approaches is generalized over a large environment which has variable situations. It does not need retraining when a new resource allocation scenario is encountered as well as recomputing when a new scenario is encountered by changes in situations. Previous studies [31, 32, 33, 34, 35], usually consider only a single scenario when they are determining optimal resource allocation. ARM control tasks are usually nonconvex optimization problems or NP-hard in many instances [36, 37]. So, the DRL concept is suitable in this domain due to its adaptability, computational efficiency, and the complexity of DRL inference. Integer programming [38] and dynamic programming [39] are mathematical models that are often used to find the exact approach to resource allocation [40], which is an NP-hard problem. However, if the practical projects under study are large or complex, these methods may not be computationally feasible or may result in a "combinatorial explosion" problem [41, 42]. These methods include [43, 44, 45, 46] which use priority rules reflecting multiple variables, including the critical index of the activity, the duration, and the minimum late finish time. In spite of this, there are few other heuristic rules that consistently perform better than all others [47]. However, it would be no basis for choosing one rule over another. It is also possible to become trapped within local optima using the general heuristic methods [48]. Tabu search (TS), simulation annealing (SA), and genetic algorithms (GA) are the three methods of metaheuristics. Repetitive improvements on current solutions are used in SA to achieve better solutions. SA has been applied for resource allocation by [43, 48, 49]. As iterations progress, TS improves the feasible solution so that a local optimum trap it to reach a global one. There are several papers using it to reexamine resource allocation, including [48, 50, 51]. GA has been applied to perform resource allocation based on evolutionary and genetic mechanisms [48, 52, 53, 54]. However, Reinforcement Learning (RL) which is similar to how learning occurs in nature is an area of machine learning. Taking action depends on the outcomes derived from previous actions by an entity called an agent. A positive reinforcement or encouragement of the behavior would reinforce or increase the importance of the action and its actions leading up to it and vice versa (see Fig. 2). The RL approach is based on Markov Decision Processes [55] and differs from supervised learning because it does not require labelled inputs-outputs [56]. In construction projects, the states are unique, and supervised machine learning is impossible since there is no dataset of actions or consequences. There is one type of ARM on which all mentioned approaches focus, and the agent does not attempt to combine information of portfolio and individual projects. On the other hand, compression of portfolio information reduces computational complexity, but the approach is not RL-based and does not take into account individual projects' information with high precision and hard resource constraints. Overall, none of the previous studies utilized the methods of parallel processing portfolio and individual projects information so that can be applied to various types of resources for ARM.

Consequently, this paper makes the following main con-

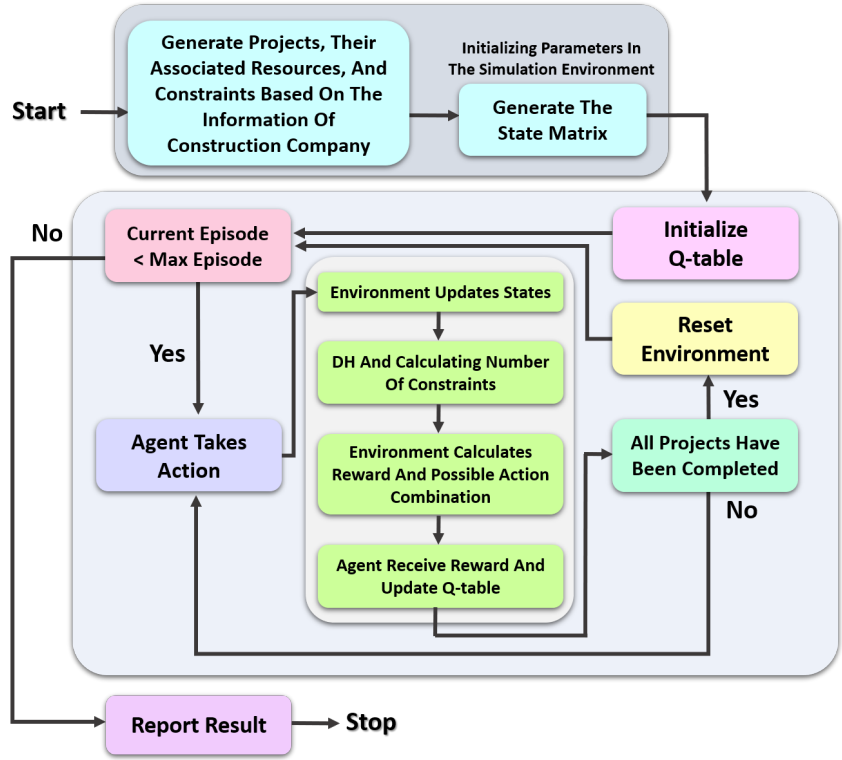

**Figure 2:** Q-Learning workflow for resource allocation in the construction company.

tributions:

- The proposed DRL method was established as a general technique for ARM by indicating its applicability to two distinct situations regarding resources: resource allocation for wireless data harvesting and coverage resource allocation;

- A novel method of exploiting the information of portfolio and individual projects is introduced for construction company that efficiently scales to large realistic projects;

- Control policy is generalized over the objectives' parameters to cover randomly generated project's resource size in order to overcome the limitations of fixed project's resource size.

- The impacts of key processing parameters of portfolio information's simulation maps on the performance of the resource allocation were investigated.

## 2. Methodology

### 2.1. Problem Formulation: Construction Company Simulation and ARM Model

In the following, it is shown how two parts of the problem are separated: the construction portfolio information and the individual projects' target, which can make the problem description universal.

Considering a square grid of size $C \times C \in \mathbb{N}^2$ with portfolio of size $m$, where $\mathbb{N}$ is the set of natural numbers. Individual projects' start/finish boundaries, constraints of resources, and other obstacles such as delay in DH make up the construction company simulation. It can be described by a tensor $\mathbf{C} \in \mathbb{R}^{C \times C \times 3}$, in which $\mathbb{R} = \{0, 1\}$ and start/finishing individual projects make up project-layer 1, constraints of resources and obstacles make up project-layer 2, and obstacles alone make up project-layer 3.

Constant individual projects $k$ is maintained by the ARM through this construction company simulation as it occupies individual projects. Therefore, its project can be defined in terms of $\mathbf{p}(t) \in \mathbb{N}^2$. A collision-avoidance strategy and constraints on resources restrict the selection of individual project by the ARM. The ARM must start and end its tasks without exceeding its minimum delay time determined by its initial resource level in any project that belongs to the start and finish projects. Every time a step in the action is completed and set to $r_0 \in \mathbb{N}$ at time $t = 0$, the resource level of the project $r(t)$ decrements by 1.

### 2.2. Target Project and Objectives Definitions

1) CRA: When allocating resources for coverage, the objectives is to minimize delay time or project time, such that falls inside the information field of a construction sensor installed on the individual project. $TP(t) \in \mathbb{R}^{C \times C}$ can describe the target project, where each resource describes whether a project needs to be covered. Each project can be classified as belonging to the current field of information by using $\mathbf{V}(t) \in \mathbb{R}^{C \times C}$ to indicate whether it belongs to it. This work uses a simulation field of information with a 5x5 size adjacent the current position of the ARM. The V(t) calculation also takes into account obstructions of Line-of-Sight. This prevents the ARM from looking around the corner.

As a result, the target project develops in accordance with:

$$\mathbf{T}P(t+1) = \mathbf{T}P(t) \wedge \neg \mathbf{V}(t) \quad (1)$$

There are two types of logical operators ∧ and ¬ which represent logical "and" and "negations". Starting and finishing points of the projects as well as constraints of resources can be covered by targets, whether stopped projects are in the construction company simulation or not. We try to complete as much of the target project within the minimum time delay as possible.

2) DH : In contrast, wireless data harvesting is utilized to obtain information of residual resources from $M \in \mathbb{N}$ terminal IoT devices scattered across the construction portfolio at simulation, with $m \in [1, M]$ situation given by $\mathbf{u}_m \in \mathbb{N}^2$. The ARM must collect a certain amount of data $D_m(t) \in I$ from each device. According to the usual log-resource loss model with Gaussian shadow fading, over the selected device $m$, their data throughout $C_m(t)$ is determined, whether the ARM including stopped projects or not. One device at a time communicates with the ARM, which is selected regarding the most available data resources and the highest resource requirements. In [15], the performance of the link and multiple access protocols are described in more details. In each device, data evolves in accordance with

$$D_m(t+1) = D_m(t) - C_m(t) \quad (2)$$

Every project can have a device. Data harvesting must collect all the data from the devices within the shortest possible time frame and minimum delay in order to collect as much data as possible.

3) Unifying Construction Project-Layer Description: $\mathbf{D}(t) \in I^{C \times C}$ is target project-layer which can be used to describe both problems. According to (1), CRA provides the target project-layer through T(t). Target project-layers in DH show how much data is available in each project in which one of the devices is located, so that project $\mathbf{u}_m$ has value $D_m(t)$ and is evolving according to (2). The value of a project is 0 if a device does not exist in the project or 1 if all the device data have been collected. Deep reinforcement learning based on a neural network with the same structure can solve both problems since their state representations are similar.

### 2.3. RL Formulation

The following approach can be applied to both distinct allocations of resources problems, despite the fact that there are a variety of methods for solving them separately. Typically, individual projects are connected by resources' limitations into a graph, and each project is covered by an allocation of resource in classical CRA approaches. As a result, CRA becomes an instance of the travelling salesman problem (TSP), which can be handled by standard approaches, such as [57], but at the expense of exponentially increasing time complexity. IoT devices can serve as nodes in a graph and distances between them as edge costs in a TSP. However, the conversion doesn't take into account communications between the device and the ARM while traveling to or from the device. A sequential visit of all devices isn't usually the optimal behavior in DH problems. Instead, information can already be obtained more efficiently by building a LoS link farther away, or the ARM may have to hover near a device for an extended period to collect information in large quantities. It is not trivial to model and solve these constraints with classical techniques, coupled with stochastic communication channel models and multiple access protocol options. It is often not possible to cover or collect all the data for both problems due to the constraints of ARM and resources. With the DRL methodology, we can directly combine all of the resource allocation goals and constraints without approximations.

## 3. Partially Observable Markov Decision Processes (POMDPs)

POMDPs [58, 59, 60, 61] are partially observable Markov decision processes [61, 62] that are defined through tuples $(S, \mathcal{A}, P, R, \Omega, \mathcal{O}, \gamma)$ used to address the problems outlined above. As shown in the POMDP, $S$ and $\mathcal{A}$ represent state space and action space, respectively, and $P : S \times S \times \mathcal{A} \mapsto \mathbb{R}$ represents transition probability function. Resources, actions, and next states to a real-valued reward are all included in the reward function $R : S \times S \times \mathcal{A} \mapsto \mathbb{R}$. A definition of observation resources is provided by $\Omega$, and an explanation of observation functions is provided by $\mathcal{O} : S \mapsto \Omega$. Long-term and short-term resource allocation rewards are valued differently according to discount factor $\gamma \in [0, 1]$

We unify the ARM problems by showing their resource state with

$$S = \underbrace{\mathbb{R}^{C \times C \times 3}}_{\substack{\text{Construction Company} \\ \text{Portfolio Simulation}}} \times \underbrace{I^{M \times M}}_{\substack{\text{Target} \\ \text{Project}}} \times \underbrace{\mathbb{N}^2}_{\text{Situation}} \times \underbrace{\mathbb{N}}_{\substack{\text{Time} \\ \text{Delay}}}, \quad (3)$$

where the components $s(t) \in S$ are

$$s(t) = (\mathbf{C}, \mathbf{D}(t), \mathbf{p}(t), r(t)) \quad (4)$$

A tuple consists of four components:

- **C** the construction company portfolio simulation containing expected start and end point of the projects, constraints of resources, and stopped projects;

- **D**(t) project indicating whether there is any more data at device on construction portfolio simulation or whether any more individual projects need to be discovered at time $t$;

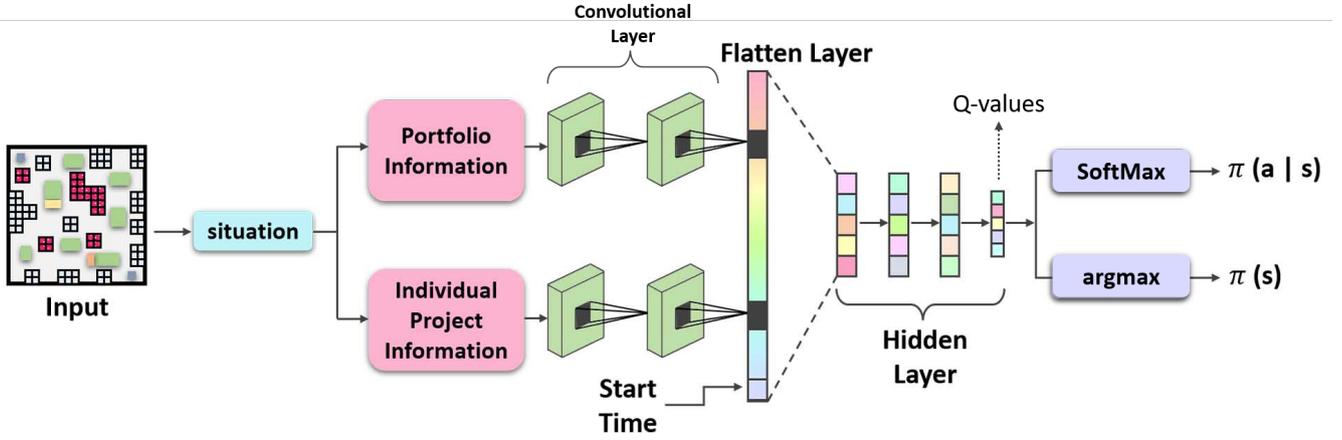

**Figure 3**: Portfolio and Individual Project Information Simulation Mapping with DQN architecture.

- $\mathbf{p}(t)$ the ARM state at time $t$;
- $r(t)$ the ARM remaining decision of individual project at time $t$;

Action $a(t) \in \mathcal{A}$ of the ARM at time $t$ is listed as one of the alternative actions

$$\mathcal{A} = \{ \text{next, previous, assign, ignore, hold} \}$$

There are four elements that make up the generalized reward function $R(s(t), a(t), s(t+1))$ :

- $r_c$ (positive) used to cover the reward derived from the collected information or the number of new target projects covered, comparing $s(t+1)$ and $s(t)$.
- $r_{sc}$ (negative) When an ARM must avoid colliding with a project or constraints of resources, safety controller penalty will be imposed.
- $r_{dec}$ (negative) When the ARM does not complete the tasks, a decision penalty is applied.
- $r_{delay}$ (negative) penalty for residual resource time delay gets zero due to safely starting the individual projects.

### 3.1. Construction Company Portfolio Simulation Processing

It is necessary to use two simulation map processing steps in order to facilitate the interpretation of the large resource state presented in Eq. (3). Also, for the purpose of improving the agent's performance, the simulation mapping should be centered around its position. By using this approach, the resource state representation size is further increased, which is a disadvantage. Moreover, the centered simulation map is presented as two inputs: detailed individual project information, showing the agent's unreliable surroundings, and squeezed portfolio information, showing the whole portfolio. This is how the three functions are mathematically described, which compose the main contribution of this work. An illustration of the data pipeline can be found in Fig. 3.

Given a tensor $\mathbf{A} \in I^{C \times C \times n}$ describing the layers of the construction company portfolio, a centered tensor $\mathbf{B} \in I^{C_m \times C_m \times n}$ with $C_m = 2C - 1$ is defined as follows:

$$\mathbf{B} = f_{\text{center}}\left(\mathbf{A}, \mathbf{p}, \mathbf{x}_{\text{pad}}\right) \quad (5)$$

According to the following definition of the centering function:

$$f_{\text{center}} : I^{C \times C \times n} \times \mathbb{N}^2 \times I^n \mapsto I^{C_m \times C_m \times n} \quad (6)$$

The components of $\mathbf{B}$ according to the components of $\mathbf{A}$ are defined as:

$$\mathbf{b}_{i,j} = \begin{cases} \mathbf{a}_{i+p_0-C+1, j+p_1-C+1}, & C \leq i + p_0 + 1 < 2C \\ & \wedge C \leq j + p_1 + 1 < 2C \\ \mathbf{x}_{\text{pad}}, & \text{otherwise} \end{cases} \quad (7)$$

The padding value $\mathbf{x}_{\text{pad}}$ is effectively applied to the project-layers of construction company portfolio of $\mathbf{A}$. There is a vector value of dimension $I^n$ in $\mathbf{a}_{i,j}$, $\mathbf{b}_{i,j}$, and a vector value of dimension $\mathbf{x}_{\text{pad}}$. There are two problems, namely constraints and stopped projects. To solve them, the construction layers are padded with $[0, 1, 1, 0]^T$. Centering is qualitatively explained in [15] and illustrated with an example. Portfolio and individual project information simulation mapping cannot be performed without using the tensor $\mathbf{B} \in I^{C_m \times C_m \times n}$, which is the result of the simulation map centering function. At the initial step, the simulation map of an individual project must be created according to:

$$\mathbf{X} = f_{\text{project}}(\mathbf{B}, proj) \quad (8)$$

by using the individual project simulation map function defined by:

$$f_{\text{project}} : I^{C_m \times C_m \times n} \times \mathbb{N} \mapsto I^{proj \times proj \times n} \quad (9)$$

When $X$ is compared to $B$, the following elements are presented:

$$\mathbf{x}_{i,j} = \mathbf{b}_{i+C-\lfloor \frac{l}{2} \rfloor, j+C-\lceil \frac{l}{2} \rceil} \quad (10)$$

As a result of this operation, a size $l \times l$ central crop has been obtained.

In order to create the portfolio simulation map, the following criteria are used:

$$\mathbf{Y} = f_{\text{portfolio}}(\mathbf{B}, port) \quad (11)$$

In order to create the portfolio simulation map, the following criteria are used:

$$f_{\text{portfolio}} : I^{C_M \times C_M \times n} \times \mathbb{N} \mapsto \mathbb{R}^{\left\lfloor \frac{C_m}{\text{port}} \right\rfloor \times \left\lfloor \frac{C_m}{\text{port}} \right\rfloor \times n} \quad (12)$$

$\mathbf{Y}$ is composed of the following elements when compared to $\mathbf{B}$:

$$\mathbf{y}_{i,j} = \frac{1}{port^2} \sum_{u=0}^{port-1} \sum_{v=0}^{port-1} \mathbf{b}_{port,i+u,port,j+v} \quad (13)$$

which is an operation equal to the average pooling.

By using $l$ and $g$, respectively, we can parameterize the functions $f_{\text{project}}$ and $f_{\text{portfolio}}$. Project simulation maps are larger when $l$ is increased, whereas the average pooling of individual projects is larger when $g$ is increased, resulting in portfolio simulation maps being smaller.

### 3.2. Observation Construction Company Portfolio Simulation

$\Omega$ which is the observation company portfolio simulation for the agent, is given as follows:

$$\Omega = \Omega_l \times \mathbb{N} \times \Omega_g$$

When $\Omega_l = \mathbb{R}^{\text{proj} \times \text{proj} \times 3} \times I^{\text{proj} \times proj}$ and $\Omega_g = I^{\left\lfloor \frac{C_m}{\text{port}} \right\rfloor \times \left\lfloor \frac{C_m}{\text{port}} \right\rfloor \times 3} \times I^{\left\lfloor \frac{C_m}{\text{port}} \right\rfloor \times \left\lfloor \frac{C_m}{\text{port}} \right\rfloor}$ are applied to the project and portfolio simulation map, respectively, and the project-layers are compressed using the average pooling, the company portfolio layers become real instead of boolean. By using the tuple, we define the observations $o(t) \in \Omega$:

$$o(t) = \left( \mathbf{C}_{proj}(t), \mathbf{D}_{\text{proj}}(t), \mathbf{C}_{port}(t), \mathbf{D}_{port}(t), r(t) \right) \quad (14)$$

A project and portfolio observation of the company portfolio simulation is represented by $\mathbf{C}_l(t)$ and $\mathbf{C}_g(t)$ in the observation. A project observation is $\mathrm{D}_l(t)$, and a portfolio observation is $\mathrm{D}_a(t)$. There is a similar time delay $r(t)$ for the ARM in the resource state. Because the ARM moves along a time-dependent path, both the project and portfolio simulation observations are timedependent. $O : S \mapsto \Omega$ describes how state of resources are mapped to observation resources, with $o(t) \in O$ defined as:

$$\begin{aligned} \mathbf{C}_{\text{proj}}(t) &= f_{\text{project}} \left( f_{\text{center}} \left( \mathbf{C}, \mathbf{p}(t), [0,1,1]^T \right), \text{ proj} \right) \\ \mathbf{D}_{\text{proj}}(t) &= f_{\text{project}} \left( f_{\text{center}} (\mathbf{D}(t), \mathbf{p}(t), 0), proj \right) \\ \mathbf{C}_{\text{port}}(t) &= f_{\text{portfolio}} \left( f_{\text{center}} \left( \mathbf{C}, \mathbf{p}(t), [0,1,1]^T \right), \text{ port} \right) \\ \mathbf{D}_{\text{port}}(t) &= f_{\text{portfolio}} \left( f_{\text{center}} (\mathbf{D}(t), \mathbf{p}(t), 0), \text{ port} \right) \end{aligned} \quad (15)$$

In this case, the issue is intentionally transformed into a partially visible MDP resource $\Omega$ by supplying it to the agent instead of the resource state $S$. Project simulation maps are restricted in size, and portfolio simulation maps are averaged. Hence, partial observability results show that partial reliability does not exacerbate the problem unsolvable for memoryless agents, as well as that the neural network becomes significantly smaller after compression, resulting in substantially less training time.

### 3.3. Double Deep Reinforcement Learning - Neural Network

A reinforcement learning approach can be used to solve the POMDP outlined above, specifically double deep Q-networks (DDQNs) [37]. According to DDQNs, each pair of state-action values is approximated as follows:

$$Q^{\pi}(s(t), a(t)) = \mathbb{E}_{\pi} \left[ \sum_{k=t}^{T} \gamma^{k-t} I(s(k), a(k), s(k+1)) \right] \quad (16)$$

An agent following policy $\pi$ will receive a discounted cumulative reward. A replay memory $\mathcal{D}$ stores $(s, a, r, s')$ the experiences $(s(t), a(t), i(t), s(t+1))$ that the agent collects as it explores the construction company simulation, omitting temporal information to converge to the optimal Q-value. $\theta$ and $\bar{\theta}$ are used to parametrize two Q-networks. $\theta$ updates the first Q-network by reducing loss:

$$L(\theta) = \mathbb{E}_{s,a,s' \sim \mathcal{D}} \left[ \left( Q_{\theta}(s,a) - Y(s,a,s') \right)^2 \right] \quad (17)$$

The replay memories are based on experiences. Target value is as follows:

$$Y(s, a, s') = r(s, a) + \gamma Q_{\bar{\theta}} \left( s', \underset{a'}{\operatorname{argmax}} Q_{\theta}(s', a') \right) \quad (18)$$

A soft update parameter $\tau \in (0, 1]$ is applied to the parameters of the second Q-network as $\bar{\theta} \leftarrow (1-\tau)\bar{\theta} + \tau\theta$. As a means of addressing training sensitivity to replay memory size, Zhang and Sutton [63] propose combined experience replay.

Fig. 3 depicts how both Q-networks are designed using neural networks. Project and portfolio observation components are generated by stacking and centering the target resource assignment and construction company simulation around the ARM. The constructed tensors are put into two branch of convolutional layers which then flatten and concatenate with the remaining time to the start point of individual projects before passing through three hidden layers using Rectified Linear Unit (ReLU) activation functions [64]. Employing a SoftMax function for exploration or argmax function for exploitation, a layer without an activation function represents the Q-values directly. To determine scalability, the flatten layer should be of a size that provides adequate scalability. It is calculated by the following equation:

$$N = n_c \left( \left( \text{proj} - n_k \left\lfloor \frac{S_k}{2} \right\rfloor \right)^2 + \left( \left\lfloor \frac{C_m}{\text{port}} \right\rfloor - n_k \left\lfloor \frac{S_k}{2} \right\rfloor \right)^2 \right) + 1$$

Table 1
Hyperparameters for two different construction company simulation.

| Description | 32*32 | 50*50 | Parameters |
| --- | --- | --- | --- |
| Trainable weight | 1,176,302 | 979,694 | $\|\theta\|$ |
| Project construction company simulation size | 17 | 17 | $proj$ |
| Portfolio construction company simulation scaling | 3 | 5 | $port$ |
| Number of convolution layers | 2 | 2 | $n_k$ |
| Number of kernels | 16 | 16 | $n_c$ |
| Convolution kernel width | 5 | 5 | $s_k$ |

Table 2
Legends for DH CRA plots.

| Mode | Description | Symbol |
| --- | --- | --- |
| DQN Input | Start and finishing point |  |
| | Stopped projects |  |
| | Projects' loss wireless links |  |
| | IoT device |  |
| | CRA: Remaining target project |  |
| Visualization | DH: Summation of resource |  |
| | DH: Updating while communicating with green device |  |
| | DH: Deciding while communicating with green device |  |
| | CRA: Not covered and covered resource |  |
| | Actions without communication |  |

$$\text{(19)}$$

with $n_c$ representing the number of kernels, $n_k$ representing the number of convolutional layers, and $s_k$ representing the number of kernel sizes. A portfolio-project construction company simulation processing scenario with $port = 1$ scaling parameter and $proj = 0$ size parameter means that there will be no down-sampling and no additional project simulation map. In Table 1, the used parameters for evaluation were listed.

### 3.4. Simulation Setup

The ARM decides in two different construction company simulation. In Fig. 4 (a and c), 'Site32' has a grid of 32 by 32 of projects with two starting and finishing points on the top left and bottom right corners. Stopped projects are also presented along with regular construction site patterns. There are a start point and a finish point around the target project in the 'Site50' scenario Fig. 4 (b and d). Stopped projects are more prominent at the bottom of the simulation map since they are generally larger and more spaced out. As the 'Site50' simulation map shows, it includes roughly larger projects. The legend for the plots provided by Table 2 shows the project sizes for the scenarios.

CRA: In this process, project size with different start and finish times and resource types are randomly sampled and layered, creating partially connected target projects. As an evaluation metric, the allocation of resources is traditionally used for CRA. This metric does not provide meaningful comparisons unless full coverage can be achieved. In this work, time delay which is constraining CRA is explored since full coverage is rarely feasible. Therefore, coverage ratio (CR) and coverage ratio and resources (CRAR) are defined as the evaluation metrics, which are both equally weighted if the ARM achieved success in resource allocation and zero otherwise. The CRAR metric is beneficial due to combining both goals, namely achieving high coverage and considering time constraints which impact target resources. Using random generated target projects as a baseline enables comparisons of performance over changing scenarios.

### 3.5. Data Harvesting

DH involves the ARM determining communication levels with devices by construction company simulation constant individual projects. Data rate is calculated using the same communications' channel parameters such as [15] for distance, random shadow fading, and LoS conditions. Similar to CRA, size of the resources is not a relevant metric. Because of randomly changing locations, data amounts, and minimum time delays of IoT devices, all data cannot be collected in all scenarios. As a result, CR is used as the evaluation metric to describes the ratio of the obtained and ac-cessible information, as soon as the information is gathered across all devices. Based on the CRA, the full information set and resource allocation performance are shown in one normalized metric by the CRAR in this context.

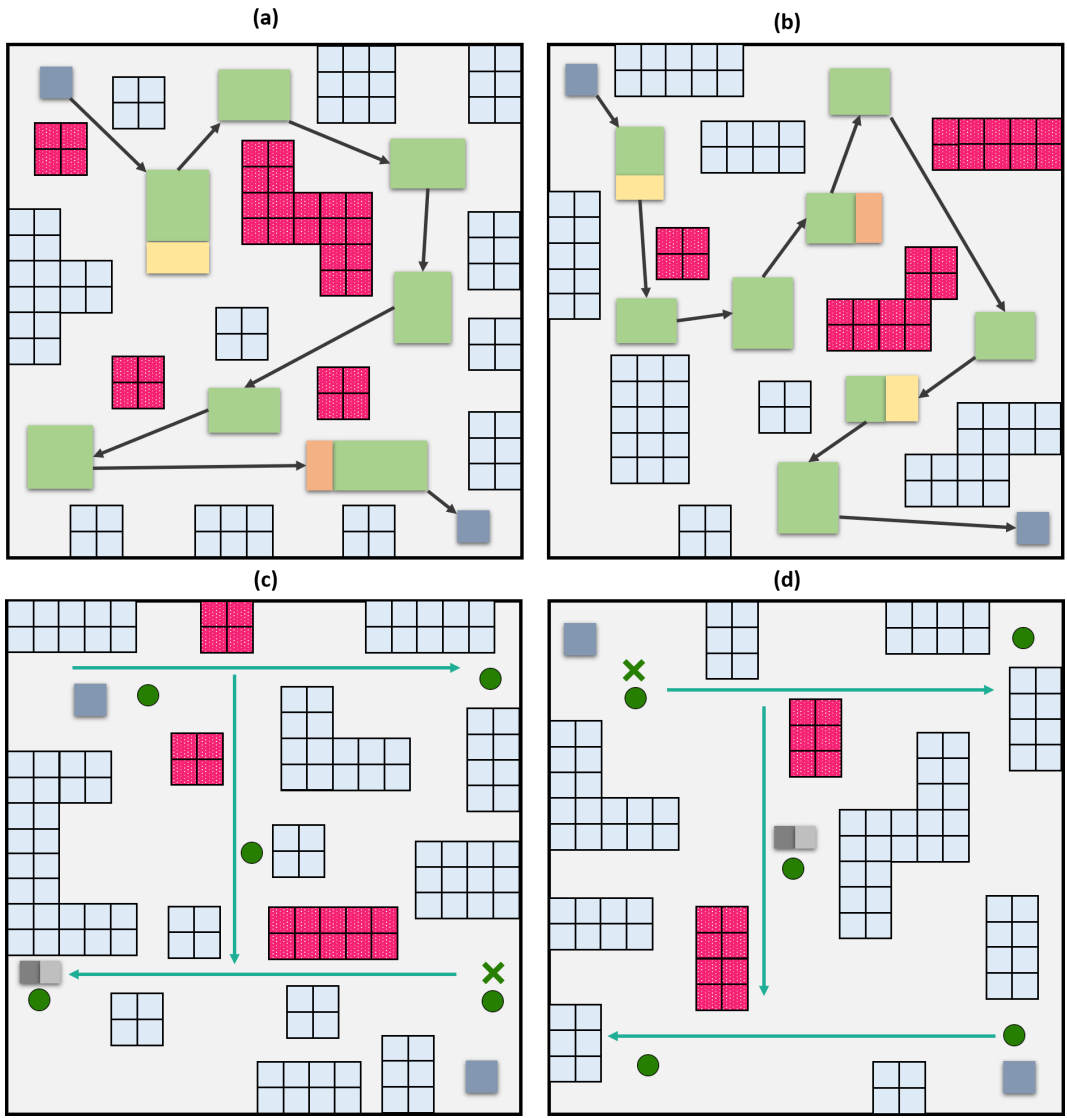

**Figure 4:** Examples of resource allocation from the Monte Carlo simulations for CRA (a)+(b) and DH (c)+(d) on "Site32" and "Site50".

## 4. General Evaluation

Approximately 20-50% of the available projects were covered by CRA agents trained on target projects containing 3-8 resources. A 'Site32' decision range of 50-150 steps was used, while a 'Site50' decision range of 150-250 steps was used. A DH scenario involves placing 3-10 devices in the project at random location. Each device contains 5.0-20.0 data units. A detailed analysis of four scenarios is presented. According to Fig. 4 (a and b), the agents in the CRA scenarios are able to find most of the resources they need to complete the target project. There is some coverage for projects which do not require much resources. The coverage of small projects is incomplete since detours are avoided. While most of the target projects are efficiently covered, there are a few exceptions. Fig. 4 (c and d) shows excellent performance from the agents in the DH scenarios. This scenario results in a collection ratio of 99.1% as small bits of information are left at the green and gray devices by the agent. It uses only 92 out of 150 possible decisions to find a concise resource. 'Site50' demonstrates that the agent collected all the data and returned them by some decision steps left over. Two million steps were completed by all four agents. According to the results of a Monte Carlo simulation using 1000 Monte Carlo scenarios (Table 3), the allocation of resources performed by all four agents is good, but that of the Site50 agent is marginally better.

A representation of the autonomous resource allocation algorithm and scheduling optimized for the projects of a construction contractor is provided in Fig. 5. The Gantt chart is depicted for a construction portfolio containing several tasks of a number of ongoing projects. In fact, the company has six projects in its portfolio, including one that has been stopped completely. These projects are different in size, tasks, and required resources at the same time. Also, a number of tasks, including civil, electrical, and mechanical, are running in parallel while two have been stopped. In addition, three types of resources are available depending on the task type, namely human resource, material, and equipment. Based on the duration and immediate predecessors and successors, the determined tasks are scheduled, and Early Start (ES), Late Finish (LF), and float are defined. The amounts of available resources are reported daily through DH, collected from IoT across all projects. Then, the resources are allocated to active tasks considering their reported requests for different resources. This process is performed in a way that the projects would be completed in the shortest possible time without or with minimum delays and penalties. Furthermore, regarding the optimum resource utilization is one of the objectives of this model. That is to say, the selected project (s) for resource allocation every day should be the one(s) that use(s) as many available resources as possible, which helps with resource-leveling. Overall, Fig. 5 shows only an example of what ARM model can do. Actually, this model can also be employed in larger or smaller companies. Moreover, the model can be utilized in every stage of the project, both from the beginning and the middle of the project as can be seen in Fig. 5.

| Project | Task | Resource | R-Rqst | Schedule ES | LF | D | F | Status | Timeline (day) 1 | 2 | 3 | 4 | 5 | 6 | 7 | 8 | 9 | 10 | 11 |
|---|---|---|---|---|---|---|---|---|---|---|---|---|---|---|---|---|---|---|---|
| 1 | Mechanical-Task1 | Human resource | 4 | 1 | 8 | 4 | 4 | Active | | | | | | | | | ✗ | | |
| | | Material | 40 | | | | | | | | | | | | | | ✗ | | |
| | Electrical-Task3 | Human resource | 2 | 2 | 5 | 3 | 1 | Active | | | | | ✓ | | | | | | |
| | | Material | 10 | | | | | | | | | | ✓ | | | | | | |
| 2 | Electrical-Task5 | - | - | 2 | 8 | 5 | 2 | Stopped | | | | | | | | | | | |
| | Civil-Task7 | Human resource | 9 | 1 | 6 | 4 | 2 | Active | | | | | ✓ | ✓ | | | | | |
| | | Material | 80 | | | | | | | | | | ✓ | ✓ | | | | | |
| | | Equipment | 3 | | | | | | | | | | ✓ | ✓ | | | | | |
| 3 | Mechanical-Task3 | Human resource | 3 | 1 | 7 | 6 | 1 | Active | | | | | | | | | | | |
| | | Material | 45 | | | | | | | | | | | | | | | | |
| | Electrical-Task4 | Human resource | 3 | 2 | 6 | 3 | 2 | Active | | | | | | | ✓ | | | | |
| | | Material | 20 | | | | | | | | | | | | ✓ | | | | |
| | Civil-Task10 | Human resource | 6 | 2 | 8 | 4 | 3 | Active | | | | | | | | | | | |
| | | Material | 90 | | | | | | | | | | | | | | | | |
| | | Equipment | 3 | | | | | | | | | | | | | | | | |
| 4 | Mechanical-Task2 | Human resource | 5 | 2 | 6 | 4 | 1 | Active | | | | | | | ✗ | ✗ | | | |
| | | Material | 55 | | | | | | | | | | | | ✗ | ✗ | | | |
| | Civil-Task1 | - | - | 1 | 5 | 3 | 1 | Stopped | | | | | | | | | | | |
| 5 | Electrical-Task5 | Human resource | 4 | 3 | 8 | 4 | 2 | Active | | | | | | | | | | | |
| | | Material | 35 | | | | | | | | | | | | | | | | |
| | Civil-Task10 | Human resource | 7 | 1 | 7 | 6 | 1 | Active | | | | | | | | ✗ | ✗ | ✗ | ✗ |
| | | Material | 80 | | | | | | | | | | | | | ✗ | ✗ | ✗ | ✗ |
| | | Equipment | 2 | | | | | | | | | | | | | ✗ | ✗ | ✗ | ✗ |
| 6 | Civil-Task3 | - | - | 2 | 6 | 3 | 2 | Stopped | | | | | | | | | | | |
| Legend | | Daily Available Resoueces: | | | | | | | | | | | | | | | | | | |
| R-Rqst | Resource Request | Human resource- Mechanical | | | | | | | 10 | 10 | 8 | 9 | 7 | 5 | 10 | 12 | 11 | 12 | 9 |
| ES | Early Start | Material- Mechanical | | | | | | | 100 | 100 | 90 | 100 | 70 | 50 | 90 | 70 | 70 | 70 | 50 |
| LF | Late Finish | Human resource- Electrical | | | | | | | 0 | 5 | 6 | 5 | 3 | 4 | 5 | 6 | 7 | 9 | 6 |
| D | Duration | Material- Electrical | | | | | | | 0 | 30 | 60 | 60 | 60 | 70 | 70 | 60 | 50 | 40 | 30 |
| F | Float | Human resource - Civil | | | | | | | 18 | 17 | 17 | 19 | 16 | 20 | 19 | 19 | 18 | 19 | 8 |
| ✓ | Earlier Finished | Material - Civil | | | | | | | 250 | 250 | 180 | 180 | 170 | 150 | 150 | 140 | 100 | 100 | 90 |
| ✗ | Delayed | Equipment | | | | | | | 5 | 5 | 5 | 3 | 4 | 4 | 4 | 4 | 3 | 3 | 3 |

**Figure 5:** Autonomous resource allocation and optimized scheduling for a sample scenario in projects of a construction company.

**Table 3**
Averaging of random scenario Monte Carlo simulations over 1000 iterations was used to determine performance metrics.

| Metric | Site32: CRA | Site32: DH | Site50: CRA | Site50: DH |
|---|---|---|---|---|
| Due time | 98.6% | 98.4% | 98.2% | 99.4% |
| CR | 71.8% | 85.6% | 83.5% | 77.5% |
| CRAR | 72.3% | 83.5% | 81.1% | 77.2% |

In the Gantt chart (Fig. 5), the best duration for the portfolio was determined by the proposed RL-based model, as expected. First, there were many violations of resource constraints in this portfolio, so intervention was required to remove them. However, doing these interventions manually would not be desirable, since it is very time-consuming especially when it is required immediately by the managers in meetings. In this case, manually solving a small problem could take a lot of time. As such, it cannot be repeated several times to analyze different situations during a meeting. However, it is possible to produce similar results within seconds employing automated methods such as GA and RL. Indeed, both of them are capable of automatically generating resource allocation avoiding violating constraints. In forthcoming meetings, such algorithms can be used to instantly compute the effects of made decisions, in a short time. The GA takes a minute for computation, while the RL algorithm only takes a second. Therefore, RL and GA perform well when dealing with small computations like this portfolio (Fig. 5). The problem is when these algorithms should be adapted to more complexed portfolios. Both algorithms are expected to perform slower as the number of projects and resources increases and resources become more constrained. Nevertheless, it was found out that even when CPU time was taken into account, the RL algorithm performed significantly better than the GA, for generating solutions, in terms of computational time. In addition, the RL algorithm maintains its quality in producing optimal solutions regardless of scale or complexity, whereas the effectiveness of genetic algorithm's result reduces. As a result, an RL-based approach always outperforms a GA-based approach with regard to processing time. Based on achieved results, the proposed RL approach in this paper demonstrates its capability to support upcoming planning by allocating resources for different scenarios in real time. Therefore, it enables project managers to interactively assess the impact of different strategies and constraints on project duration and then make the optimal decision. As part of this method, the IoT and automated DH technologies are incorporated to assist project managers to plan ahead error-free resource management.

**Table 4**
Flatten layer size for a comparison between two different strategies of simulation map processing, *port* and *proj*.

| Portfolio construction company simulation scaling *port* | Project construction company simulation scaling *proj* | | | |
|---|---|---|---|---|
| | 9 | 17 | 25 | 33 |
| 2 | 8,581 | 9,751 | 13,189 | 18,455 |
| 3 | 2,751 | 4,003 | 7,339 | 12,704 |
| 5 | 274 | 1,543 | 4,882 | 10,254 |
| 7 | 33 | 1,323 | 4,631 | 10,015 |

### 4.1. Portfolio-Project Parameter Evaluation

Our study tested multiple situations by single agent with different variable on the CRA and DH problems in order to determine whether the new hyperparameters, portfolio simulation map scaling called *port*, and project simulation map size called *proj*, have a significant impact on the performance. To test each possible combination, three agents for four values of *proj* as well as four values of port were trained. Furthermore, three agents were trained without project and portfolio construction company simulation processing which would correspond to *proj* = 0 and *port* = 1. Based on 200 Monte Carlo scenarios, 51 agents were created for the CRA and DH problems. There was a difference between the previous assessment and this one because 150-300 decision steps were used. Following are the parameters selected according to (19) and the flatten layer size, as can be seen in the Table 4. The CRAR values for the CRA and DH problems are shown in Fig. 4 (a and b), respectively, for each agent's flatten layer size. The training process is significantly faster than that for agents without portfolio and project simulation map processing. Parameters are more important in the DH problem than in the CRA problem, as can be seen in the Table 5. Flattening layers are generally more effective up to a point when they have a larger thickness. CRAR can be zero for some runs for both problems because of a large flatten layer. In this case, the agent does not properly allocate resources due to a lack of learning. As DH agents don't use the portfolio-project simulation map approach, they have a very low CRAR score since they don't learn how to allocate consistently. Table 1 shows that the agents with *proj* = 17 and *port* = 5 or *port* = 3 present the best performance in both scenarios despite their small flatten layer sizes of only 33 neurons. Apart from these two parameter combinations, the agents with *port* = 7 and *proj* = 9 also perform well with these parameters.

A comparison of the time for resource allocation using various strategies is presented in Table 5, which shows that CRA and DH will be able to train with the proposed method of portfolio simulation processing faster than when no portfolio simulation processing is used. As a result, in the proposed method, there is no limitation in scheduling and allocating resources according to the number of projects. Although increasing the number of projects makes GA impractical due to the computation time, it will not change the computation time for resource allocation and project scheduling in the proposed method. Also, it can lead to non-optimal scheduling and allocation by the GA.

## 5. Conclusion

As construction industry is highly competitive, optimal resource management is one of the most important roles of the project manager. Computer application software offer the advantage of accurate resource management which distinguishes it from manual methods. The main objective of this paper is to present a general solution for ARM and resource allocation that can be applied to two distinct objectives, coverage resource allocation (CRA) and data harvesting (DH). The reward function was designed to combine specific objectives with decision constraints so that DDQNs could efficiently assign resources in both scenarios. In this paper, a novel portfolio-project simulation map processing scheme was presented that determines how simulation map parameters affect the learning performance of the DRL agent, allowing large simulation maps to be directly fed into convolutional layers.

**Table 5**
A comparison between two different strategies of simulation map processing.

| Portfolio construction company | Project construction company simulation scaling *proj* | | | | | | | |
|---|---|---|---|---|---|---|---|---|
| | 9 | | 17 | | 25 | | 33 | |
| simulation scaling *port* | CRA | DH | CRA | DH | CRA | DH | CRA | DH |
| 2 | 2.7 | 2.3 | 2.3 | 2.0 | 1.8 | 1.5 | 1.2 | 1.1 |
| 5 | 3.4 | 2.9 | 3.1 | 2.4 | 2.0 | 1.8 | 1.6 | 1.4 |
| 7 | 4.3 | 3.5 | 3.4 | 3.1 | 2.5 | 2.3 | 1.8 | 1.6 |

The proposed model is beneficial for numerous projects of different sizes and with different amounts of resources. In spite of previous methods such as the genetic algorithm, this model eliminates the computational weaknesses in terms of time and complexity. In fact, a few seconds are required to perform calculations of optimal resource allocation, resource-leveling, and scheduling of projects. Furthermore, the proposed model is designed in such a way that it can schedule the rest of the tasks from any stage of the project. Due to the very short time required for doing calculations and providing results by this model, it can be used in every phase of the project with the improvements of the model in the future. For instance, using this model in decision-making meetings enables senior managers to make the optimal decision.

The next step is to examine the remaining hindrances to the application of our method to even larger construction company simulation, namely micro-alternation of decisions through macro-actions or options. In the future, it will also be possible to perform experiments with realistic ARM simulators using the combination of the presented high-level resource allocation approach and the low-level time delay controller. The performance of resource allocation will be further examined by investigating irregular projects and non-convex obstacles.

## 6. Disclosures

The authors declare no conflict of interest.